\documentclass[10pt]{article}

\usepackage[table]{xcolor}
\usepackage{techreport}
\usepackage{amsmath}
\usepackage{graphicx}
\usepackage{booktabs}
\usepackage{multirow}
\usepackage{pifont}
\usepackage{tikz}
\usepackage{microtype}
\usepackage[numbers,sort&compress]{natbib}
\usepackage[colorlinks=true,linkcolor=accent,citecolor=accent,urlcolor=accent,bookmarks=true]{hyperref}
\usepackage{xurl}

\newcommand{\cmark}{$\checkmark$}

\definecolor{rwlabel}{RGB}{38,84,160}
\definecolor{rwact}{RGB}{122,46,160}
\definecolor{rwgoal}{RGB}{22,128,58}
\definecolor{rwbg}{RGB}{238,243,250}
\newlength{\rwgap}
\newlength{\rwwidth}
\newcommand{\rwact}[1]{\textcolor{rwact}{\textbf{#1}}}
\newcommand{\rwgoal}[1]{\textcolor{rwgoal}{\textbf{#1}}}
\newcommand{\rwinstr}[2]{%
  \fcolorbox{rwlabel!35}{rwbg}{\parbox{\dimexpr\linewidth-2\fboxsep-2\fboxrule\relax}{%
    \small\raggedright\textcolor{rwlabel}{\textbf{#1:}} #2}}}
\newcommand{\rwframe}[3]{%
  \begin{tikzpicture}
    \node[inner sep=0pt, anchor=south west] (img) at (0,0)
      {\includegraphics[width=\rwwidth]{figures/realworld/#1.jpg}};
    \node[anchor=south west, fill=black, fill opacity=0.6, text opacity=1,
      text=white, font=\scriptsize\sffamily, inner sep=1.6pt, rounded corners=1pt]
      at ([shift={(2pt,2pt)}]img.south west) {#2};
    \ifnum#3=1
      \node[anchor=south east, circle, fill=white, fill opacity=0.9, text opacity=1,
        text=rwgoal, font=\Large, inner sep=1.2pt]
        at ([shift={(-2pt,2pt)}]img.south east) {\ding{51}};
    \fi
  \end{tikzpicture}}

\renewcommand{\shorttitle}{StageVLN: Spatial and Trajectory Auxiliary Guidance for Efficient VLN}

\title{StageVLN: Spatial and Trajectory Auxiliary Guidance\\for Efficient Vision-Language Navigation}

\author{%
  \mbox{Anh Dao\textsuperscript{1,*}} \and
  \mbox{Quan-Dung Pham\textsuperscript{1,*}} \and
  \mbox{Le Danh Vinh\textsuperscript{1,*}} \and
  \mbox{The Anh Nguyen\textsuperscript{1}} \and
  \mbox{Nguyen Viet Tri Pham\textsuperscript{1}} \and
  \mbox{Yiyu Chen\textsuperscript{1}} \and
  \mbox{Pham Tuyen Le\textsuperscript{1}} \and
  \mbox{Van-Truong Nguyen\textsuperscript{1}} \and
  \mbox{Quan Nguyen\textsuperscript{2}}}
\affiliations{%
  \textsuperscript{1}VinMotion, Inc., Vietnam\qquad
  \textsuperscript{2}University of Southern California, USA\qquad
  \textsuperscript{*}Equal contribution}
\projectpage{https://stagevln.github.io}

\hypersetup{pdftitle={StageVLN: Spatial and Trajectory Auxiliary Guidance for Efficient Vision-Language Navigation}}

\begin{document}

\maketitle

\begin{abstract}
Vision-and-Language Navigation (VLN) policies increasingly benefit from strong semantic priors provided by large vision-language models (VLMs). However, standard action supervision does not explicitly encourage intermediate representations to preserve scene geometry, relative orientation, or global episode progress. Incorporating depth estimators, explicit maps, point clouds, or geometry foundation models at inference can provide such structure but introduces additional computation, memory overhead, and architectural dependence during deployment. We introduce StageVLN, a training framework that shapes navigation representations through privileged spatial and trajectory guidance while preserving the original inference pathway. A frozen geometry foundation model provides multi-level spatial guidance to hierarchical navigator states, while relative-heading and expert-route progress objectives provide complementary trajectory-state supervision. All auxiliary components are used only during training and removed at deployment. On R2R-CE validation-unseen, StageVLN achieves 56.3\% SR and 51.4\% SPL with a 4B-parameter backbone, without an additional geometry encoder at inference. On RxR-CE, it achieves 54.3\% SR without additional navigation training data or a geometry encoder at inference.

\end{abstract}

\section{Introduction}
Vision-and-Language Navigation (VLN) tasks require an agent to process natural language instructions and egocentric visual observations to generate actions that reach a goal through diverse environments. The Room-to-Room (R2R) benchmark \cite{anderson2018vision} established an instruction-grounded evaluation framework within real scanned scenes. Subsequently, Room-Across-Room (RxR) \cite{ku2020room} extended this paradigm to support greater language and trajectory diversity, while Vision-and-Language Navigation in Continuous Environments (VLN-CE) \cite{krantz2020beyond} further transitioned the task from discrete navigation graphs to continuous 3D spaces. In continuous settings, merely recognizing referenced objects is insufficient; navigation policies must also relate the visible structure of the environment to traversable motion, maintain spatial orientation across sequential observations, and continuously localize their position along the requested route.

\begin{figure}[t]
    \centering
    \includegraphics[
        width=\textwidth,
        trim=0 8 0 8,
        clip
    ]{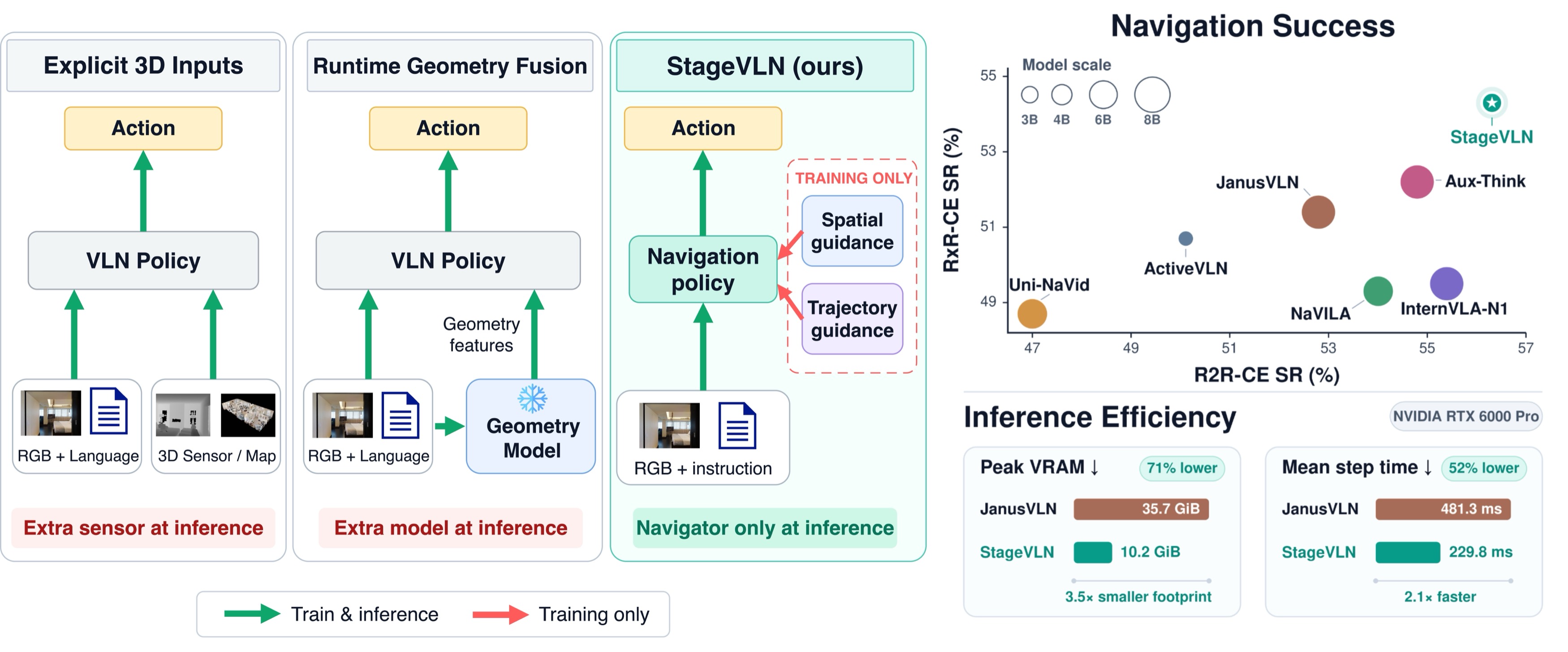}
    \caption{
    \textbf{StageVLN overview.}
    \textbf{Left:} comparison with explicit 3D inputs and runtime geometry fusion;
    StageVLN uses spatial and trajectory guidance only during training and retains
    the navigator alone at inference.
    \textbf{Right:} navigation success and inference efficiency.
    VRAM and latency are measured on a single NVIDIA RTX 6000 Pro.
    }
    \label{fig:stagevln_overview}
\end{figure}

Recent RGB-only systems \cite{zhang2024navid, zhang2024uni, cheng2024navila, wei2025streamvln, zhang2026embodied} increasingly integrate VLMs into navigation policies, leveraging images, video context, memory, or world-action modeling objectives to learn more robust behaviors. However, their strong semantic priors do not guarantee that the hidden representations used for action prediction preserve the geometric and trajectory structure required for navigation in complex environments. A direct solution is to incorporate pretrained geometry models \cite{wang2025vggt, wang2026vggt} into the inference pipeline, or to fuse depth, point clouds, maps, or latent 3D features directly into the navigation policy. While this paradigm has proven effective for spatial understanding in VLMs \cite{fan2026vlm, zhang2026spatialstack, wang2026learning} and has recently been adapted for VLN \cite{zeng2026janusvln}, it introduces extra runtime dependencies and increases deployment costs, particularly on real-world robots with constrained hardware. To address this limitation, our central question is: \textit{can geometric guidance and trajectory-state supervision improve the navigator while preserving its original deployment pathway?} This perspective aligns with the broader principles of learning via auxiliary objectives and intermediate representation distillation \cite{hinton2015distilling, lopez2015unifying}, while adapting them to the temporal structure of embodied navigation.

We propose \textbf{StageVLN}, a general framework for \textit{spatial and trajectory auxiliary guidance}. The central idea is to introduce complementary supervision at three levels of navigation structure, progressing from \textbf{local scene geometry} to \textbf{relational orientation} and finally to \textbf{global trajectory state}. During training, a frozen geometry encoder provides multi-level spatial information by supervising intermediate visual states via representation distillation. Additionally, relative heading prediction encourages the learning of a relational orientation structure across observations, while progress prediction provides the agent with a global estimate of its position along the expert trajectory. Crucially, these objectives are applied strictly at training time, allowing the navigator to retain the benefits of geometric and trajectory supervision without requiring auxiliary components at inference. As illustrated in Fig.~\ref{fig:stagevln_overview}, our contributions are threefold:
\begin{itemize}

\item We introduce a \textbf{training-only hierarchical geometry-teacher mechanism} that transfers spatial representations into multiple navigator layers without requiring geometric feature fusion at inference time.

\item We propose a \textbf{unified spatial-and-trajectory training formulation} that combines multi-level spatial feature distillation with relative-heading and expert-route progress supervision.

\item We conduct \textbf{extensive navigation evaluations and comprehensive component analyses}, demonstrating improvements in success-related metrics while characterizing localization trade-offs and distinguishing auxiliary-free deployment from measured system efficiency.

\end{itemize}

\section{Related Work}
\label{sec:related_work}

\subsection{Vision-Language Models for Navigation}
Recent VLN methods adapt pretrained vision-language models to
predict actions from egocentric observations and instructions.
NaVid~\cite{zhang2024navid} uses monocular video for end-to-end
navigation, while Uni-NaVid~\cite{zhang2024uni} extends video-based
action modeling to multiple embodied navigation tasks.
NaVILA~\cite{cheng2024navila} connects language-conditioned navigation
with legged locomotion through intermediate movement commands.
StreamVLN~\cite{wei2025streamvln} addresses long observation streams
through slow--fast context modeling, memory compression, and KV-cache
reuse. StageVLN addresses the
complementary problem of supervising the spatial and trajectory
information encoded within an RGB--language policy. It introduces
additional training objectives while preserving the base policy's
observation interface and action-generation architecture at deployment.

\subsection{Geometry-Aware Representations and Spatial Distillation}
Geometry foundation models provide representations learned from
geometric prediction tasks. VGGT~\cite{wang2025vggt}, for example,
jointly predicts camera parameters, depth, point maps, and point tracks
from images. JanusVLN~\cite{zeng2026janusvln}
combines visual-semantic and spatial-geometric features with separate
implicit memories for navigation. In spatial question answering,
SpatialStack~\cite{zhang2026spatialstack} integrates geometric features
at multiple language-model layers, demonstrating the value of
hierarchical geometry--language fusion. Both designs compute
geometric features as part of the inference pathway.

An alternative is to transfer teacher representations during training.
FitNets~\cite{romero2015fitnets} established intermediate-feature
supervision as a mechanism for knowledge transfer. More directly
related, Spatial Forcing~\cite{li2026spatial} aligns intermediate
visual embeddings of manipulation VLAs with pretrained geometric
representations while leaving their inference architecture unchanged.

\subsection{Auxiliary Supervision for Navigation}
Auxiliary objectives have long been used to improve navigation
representations beyond action imitation. The self-monitoring
agent~\cite{ma2019selfmonitoring} introduces progress estimation to
support instruction grounding. AuxRN~\cite{zhu2020auxiliary} jointly
learns trajectory retelling, progress estimation, next-orientation
prediction, and cross-modal matching; its progress objective already
uses soft step-fraction targets with cross-entropy supervision. However, these auxiliary objectives do not explicitly transfer spatial representations from a pretrained geometry model into the navigation policy.

StageVLN combines hierarchical geometric feature transfer with two trajectory-state targets. Its heading objective supervises the accumulated relative rotation between historical and current views, rather than the orientation of the next expert action. Its progress target measures the fraction of expert forward actions already completed, excluding turns, and is predicted from a pre-action hidden state without access to the supervised action token. Together with current-view geometric alignment, they provide supervision for scene structure, inter-observation orientation, and route-level state. All auxiliary components, including the geometry teacher, are removed at inference, retaining the original RGB–language navigation pathway.

\section{Method}
\subsection{Problem Formulation}

\begin{figure}[t]
    \centering
    \includegraphics[width=\textwidth]{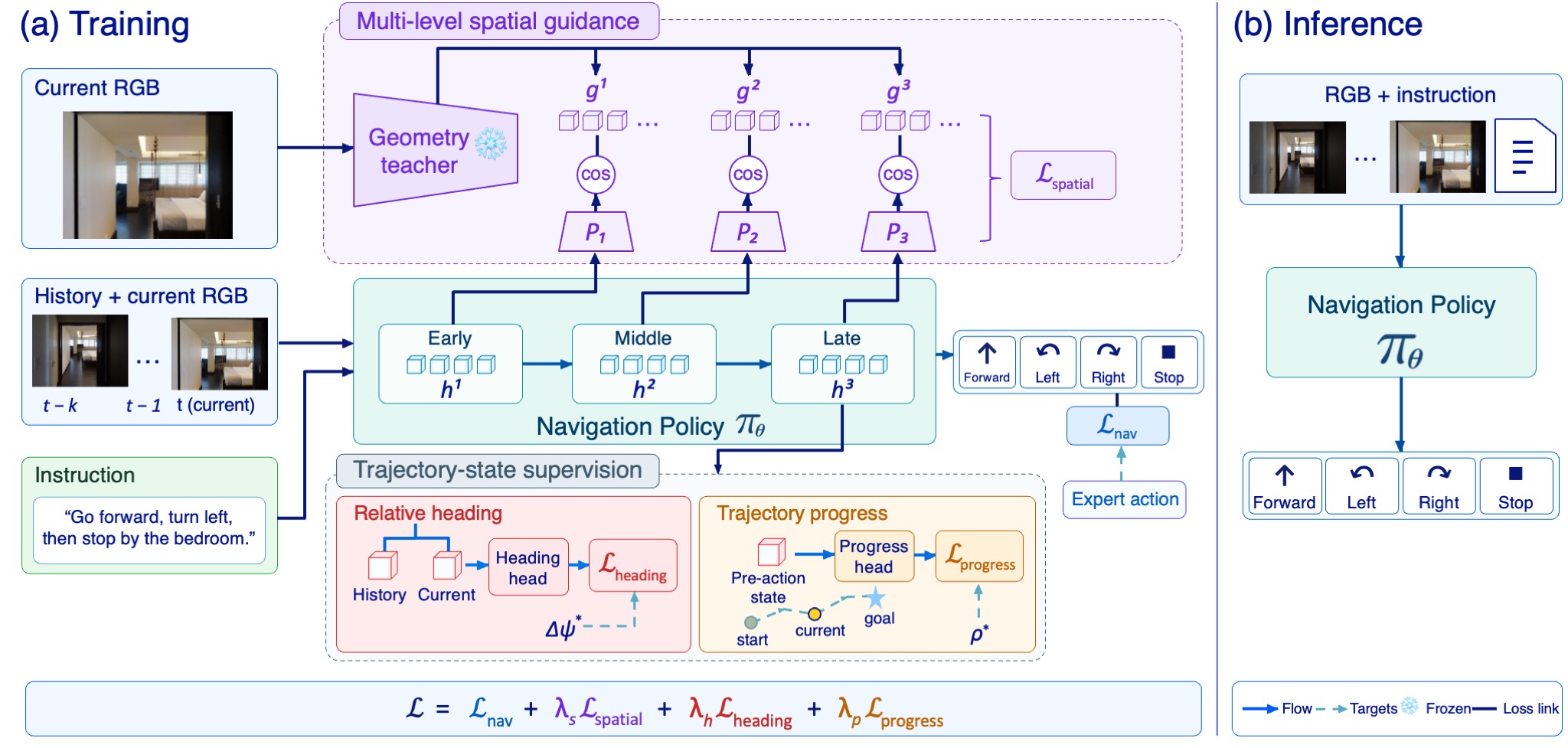}
    \caption{\textbf{StageVLN training and inference pipeline.}
    During training, a frozen geometry teacher provides multi-level
    spatial guidance, while relative-heading and expert-route progress
    objectives supervise trajectory state. At inference, the teacher,
    alignment projectors, and auxiliary heads are excluded, leaving only
    the base RGB--language navigation policy.}
    \label{fig:stagevln_method}
\end{figure}

At timestep $t$, given a language instruction $l$ and an RGB
observation history $O_t=(o_0,\ldots,o_t)$, the navigation policy
$\pi_\theta$ predicts an action $a_t\in\mathcal{A}$, where
$\mathcal{A}=\{\texttt{MOVE\_FORWARD}, \texttt{TURN\_LEFT},
 \texttt{TURN\_RIGHT}, \texttt{STOP}\}$.
Given expert demonstrations
$(l,O_t,a_t^\star)\sim\mathcal{D}$, supervised navigation minimizes
the negative log-likelihood of the expert action:
\begin{equation}
\mathcal{L}_{\mathrm{nav}}
=
-\mathbb{E}_{\mathcal{D}}
\left[
\log \pi_\theta
\left(a_t^\star \mid l,O_t\right)
\right],
\label{eq:nav_loss}
\end{equation}
where $\mathcal{D}$ denotes the expert demonstration dataset. Here, $\pi_\theta(a_t^\star\mid l,O_t)$ denotes the
autoregressive likelihood of the complete textual action label.

StageVLN additionally uses a frozen geometry teacher $g_\phi$
and expert-derived trajectory labels during training to
supervise spatial representations and trajectory state.

\subsection{StageVLN Overview}
Standard action supervision constrains policy outputs but does not
explicitly constrain intermediate representations to preserve scene
geometry or trajectory state. StageVLN introduces three complementary
auxiliary training signals: (i) hierarchical spatial representations
from a frozen geometry teacher; (ii) relative orientation between
historical and current views; and (iii) global expert-route progress. The first is
\textit{multi-level geometric representation distillation}; the latter
two constitute \textit{trajectory-state auxiliary supervision}.
Fig.~\ref{fig:stagevln_method} summarizes training and inference.

\subsection{Multi-Level Geometric Representation Distillation}

For the current observation $o_t$, a frozen pretrained geometry
teacher $g_\phi$ provides spatial features at $M$ selected layers:
\begin{equation}
\{G_t^{(r_m)}\}_{m=1}^{M} = g_\phi(o_t),
\end{equation}
where $r_m$ indexes the teacher layer used for the $m$-th
alignment pair. Following \cite{li2026spatial}, we augment
these features with positional encodings to provide explicit
spatial position information. The teacher targets are
\begin{equation}
Z_t^{(m)}
=
\operatorname{sg}\left[
R_t\!\left(G_t^{(r_m)} + E_t^{(r_m)}\right)
\right],
\label{eq:spatial_targets}
\end{equation}
where $E_t^{(r_m)}$ denotes the positional encoding and $R_t$
reshapes the teacher patch features into a two-dimensional grid,
resizes it to the navigator's current-image grid, and flattens
it in the corresponding spatial order. Non-patch tokens are
excluded. The stop-gradient operator $\operatorname{sg}$
detaches the targets, and the teacher remains frozen throughout
training.

Let $H_t^{(\ell)}$ denote the navigator's hidden sequence at
layer $\ell$. For each student--teacher layer pair
$(\ell_m,r_m)$, we extract the current-image visual tokens
$V_t^{(\ell_m)}$ from $H_t^{(\ell_m)}$. This selection excludes
historical images, language, actions, and padding. An independent
trainable projector $P_m$ maps these tokens to the teacher
feature dimension. We minimize the mean cosine distance between
the projected student features and their corresponding targets:
\begin{equation}
\mathcal{L}_{\mathrm{spatial},t}^{(m)}
=
\frac{1}{N_t}\sum_{n=1}^{N_t}
\left[
1-
\cos\!\left(
P_m(V_{t,n}^{(\ell_m)}), Z_{t,n}^{(m)}
\right)
\right],
\label{eq:spatial_level_loss}
\end{equation}
where $N_t$ is the number of aligned current-image tokens and
$\cos(\cdot,\cdot)$ denotes cosine similarity with
$\epsilon$-stabilized $\ell_2$ normalization. The spatial
objective averages over alignment levels and training samples:
\begin{equation}
\mathcal{L}_{\mathrm{spatial}}
=
\mathbb{E}_{\mathcal{D}}
\left[
\frac{1}{M}\sum_{m=1}^{M}
\mathcal{L}_{\mathrm{spatial},t}^{(m)}
\right].
\label{eq:spatial_loss}
\end{equation}
Gradients update the projectors and trainable navigator
parameters, while teacher targets remain detached.

This objective aligns selected hidden representations with
a hierarchy of geometric teacher features, rather than
matching output logits or predicting explicit geometry maps.

\subsection{Trajectory-State Auxiliary Supervision}
We complement geometric representation distillation with two trajectory-level objectives that supervise relative orientation and expert-route translation progress.

\subsubsection{Relative Heading}
Geometric guidance captures spatial structure within the current
observation, but does not explicitly supervise how the agent's
orientation changes across steps. We therefore introduce relative-heading
supervision between each selected historical observation $o_{\tau_i}$
and current observation $o_t$.

At a selected policy layer $\ell_h$, we mean-pool the visual tokens
of each frame and apply a shared frame projector $F$:
\begin{equation}
e_{\tau}
=
F\!\left(
\frac{1}{N_{\tau}}
\sum_{n=1}^{N_{\tau}}
V_{\tau,n}^{(\ell_h)}
\right),
\label{eq:heading_frame_embedding}
\end{equation}
where $V_{\tau,n}^{(\ell_h)}$ denotes the $n$-th visual token of
observation $o_{\tau}$ within the current hidden sequence
$H_t^{(\ell_h)}$, and $N_{\tau}$ is its visual-token count.
For each history-current pair, a shared relation head $R$
predicts a relative-heading vector:
\begin{equation}
u_{i,t}
=
R\!\left(
[e_{\tau_i};\,e_t;\,e_t-e_{\tau_i};\,e_{\tau_i}\odot e_t]
\right)
\in\mathbb{R}^{2},
\label{eq:heading_pred}
\end{equation}
where $[\cdot;\cdot]$ denotes concatenation and $\odot$ denotes
element-wise multiplication.

The target represents current orientation relative to history,
derived from the ground-truth expert actions:
\begin{equation}
\begin{aligned}
\Delta\psi_{i,t}
&=
\operatorname{wrap}_{[-\pi,\pi)}
\left(\sum_{s=\tau_i}^{t-1}\delta(a_s^\star)\right),\\
r_{i,t}
&=
[\sin\Delta\psi_{i,t},\,\cos\Delta\psi_{i,t}]^\top,
\end{aligned}
\label{eq:heading_target}
\end{equation}
where $\delta$ assigns positive nominal increments to left
turns, negative increments to right turns, and zero to forward
and stop actions. The current action $a_t^\star$ is excluded.
The sine–cosine encoding avoids the angular wrap discontinuity.

We minimize the heading loss:
\begin{equation}
\mathcal{L}_{\mathrm{heading}}
=
\mathbb{E}_{\mathcal{D}}
\left[
\frac{1}{|\mathcal{H}_t|}
\sum_{i\in\mathcal{H}_t}
\left(1-\cos(u_{i,t},r_{i,t})\right)
\right],
\label{eq:heading_loss}
\end{equation}
where $\mathcal{H}_t$ indexes valid selected historical frames.
This objective encourages the navigator to encode
relational orientation across observations.

\subsubsection{Expert-Route Progress}
Relative-heading supervision captures orientation changes,
while progress supervision encourages global trajectory-state
representations. For an expert trajectory containing $T$
actions, we define
\begin{equation}
p_t =
\frac{
\sum_{s=0}^{t-1}
\mathbf{1}[a_s^\star=\texttt{MOVE\_FORWARD}]
}{
\sum_{s=0}^{T-1}
\mathbf{1}[a_s^\star=\texttt{MOVE\_FORWARD}]
},
\label{eq:progress_target}
\end{equation}
where $\mathbf{1}[\cdot]$ is the indicator function.
This target measures the fraction of expert forward actions
completed before timestep $t$; turns and stopping do not
contribute.

Let $j_t$ denote the first supervised action-token position.
A progress head $Q$ predicts a logit from the immediately
preceding hidden state:
\begin{equation}
z_t = Q\!\left(H_{t,j_t-1}^{(\ell_p)}\right).
\label{eq:progress_prediction}
\end{equation}
Using this pre-action state prevents access to the ground-truth
action token under teacher forcing. Over progress samples
$\mathcal{B}_P$, we minimize binary cross-entropy with soft targets:
\begin{equation}
\mathcal{L}_{\mathrm{progress}}
=
\frac{1}{|\mathcal{B}_P|}
\sum_{t\in\mathcal{B}_P}
\left[
\operatorname{softplus}(z_t)-p_tz_t
\right].
\label{eq:progress_loss}
\end{equation}
Progress denotes expert-route completion rather than
goal-success probability; the head is excluded at inference
and provides no stopping rule.

\subsection{Joint Training Objective}
The final objective combines navigation supervision with
spatial and trajectory guidance:
\begin{equation}
\mathcal{L}
=
\mathcal{L}_{\mathrm{nav}}
+\lambda_s\mathcal{L}_{\mathrm{spatial}}
+\lambda_h\mathcal{L}_{\mathrm{heading}}
+\lambda_p\mathcal{L}_{\mathrm{progress}},
\label{eq:joint_loss}
\end{equation}
where $\lambda_s$, $\lambda_h$, and $\lambda_p$ control the
auxiliary contributions. Together, these objectives encourage
local and hierarchical scene structure, relative orientation
across observations, and global expert-route progress.

At inference, only $\pi_\theta(a_t\mid l,O_t)$ is evaluated.
The geometry teacher, alignment projectors, and trajectory
heads are excluded from the action-generation pathway,
preserving the base RGB--language policy and the representations
shaped by auxiliary guidance during training.

\section{Experiments}
\label{sec:experiments}

\subsection{Experimental Setup}

\subsubsection{Benchmarks and Training Datasets}
We evaluate StageVLN on R2R-CE and RxR-CE
\cite{anderson2018vision,ku2020room,krantz2020beyond}
in the Habitat~\cite{habitat19iccv} simulator using Matterport3D~\cite{chang2017matterport3d} indoor scenes.
Navigation fine-tuning uses only expert demonstrations from
the joint training sets: 10,819 episodes from R2R-CE and
19,990 from RxR-CE, without additional navigation datasets.
Evaluation uses the validation-unseen splits, comprising
1,839 trajectories for R2R-CE and 3,669 for RxR-CE.
We report navigation error (NE), oracle success (OS),
success rate (SR), and success weighted by path length (SPL),
with normalized dynamic time warping (nDTW) additionally
reported for RxR-CE.

\begin{table}[t]
    \centering
    \caption{Comparison on the R2R-CE and RxR-CE validation-unseen
    splits. $\downarrow$/$\uparrow$ indicate lower/higher is better.
    Best and second-best results are shown in \textbf{bold} and
    \underline{underline}, respectively.
    StreamVLN$^{*}$ uses EnvDrop augmentation, and NaVILA$^{*}$
    excludes human-following data.
    $\dagger$ denotes methods with an additional geometry encoder
    at inference; parameter counts are reported as
    VLM + geometry encoder.
    StageVLN uses no additional navigation training data beyond
    R2R-CE and RxR-CE; pretrained model data are excluded from
    the external-data counts.}
    \label{tab:stagevln_main_results}

    \setlength{\tabcolsep}{3.2pt}
    \renewcommand{\arraystretch}{1.08}

    \resizebox{\textwidth}{!}{%
    \begin{tabular}{@{}l c c c c c c c c c c c c c c@{}}
        \toprule
        \multirow{2}{*}{Method} &
        \multirow{2}{*}{\shortstack{Model\\Params}} &
        \multicolumn{4}{c}{Observation} &
        \multicolumn{4}{c}{R2R Val-Unseen} &
        \multicolumn{4}{c}{RxR Val-Unseen} &
        \multirow{2}{*}{\shortstack{External\\Training\\Data}} \\

        \cmidrule(lr){3-6}
        \cmidrule(lr){7-10}
        \cmidrule(lr){11-14}

        & &
        Pano. & Odo. & Depth & S.RGB &
        NE$\downarrow$ & OS$\uparrow$ & SR$\uparrow$ & SPL$\uparrow$ &
        NE$\downarrow$ & SR$\uparrow$ & SPL$\uparrow$ & nDTW$\uparrow$ &
        \\

        \midrule

        HPN+DN \cite{krantz2021waypoint}
        & -- & \cmark & \cmark & \cmark &
        & 6.31 & 40.0 & 36.0 & 34.0
        & -- & -- & -- & -- & -- \\

        CMA \cite{hong2022bridging}
        & -- & \cmark & \cmark & \cmark &
        & 6.20 & 52.0 & 41.0 & 36.0
        & 8.76 & 26.5 & 22.1 & 47.0 & -- \\

        Sim2Sim \cite{krantz2022sim}
        & -- & \cmark & \cmark & \cmark &
        & 6.07 & 52.0 & 43.0 & 36.0
        & -- & -- & -- & -- & -- \\

        VLN-BERT \cite{hong2020recurrent}
        & -- & \cmark & \cmark & \cmark &
        & 5.74 & 53.0 & 44.0 & 39.0
        & 8.98 & 27.0 & 22.6 & 46.7 & -- \\

        Ego2-Map \cite{hong2023learning}
        & -- & \cmark & \cmark & \cmark &
        & 5.54 & 56.0 & 47.0 & 41.0
        & -- & -- & -- & -- & -- \\

        DreamWalker \cite{wang2023dreamwalker}
        & -- & \cmark & \cmark & \cmark &
        & 5.53 & 59.0 & 49.0 & 44.0
        & -- & -- & -- & -- & -- \\

        GridMM \cite{wang2023gridmm}
        & -- & \cmark & \cmark & \cmark &
        & 5.11 & 61.0 & 49.0 & 41.0
        & -- & -- & -- & -- & -- \\

        Reborn \cite{an20221st}
        & -- & \cmark & \cmark & \cmark &
        & 5.40 & 57.0 & 50.0 & 46.0
        & 5.98 & 48.6 & 42.0 & \textbf{63.3} & -- \\

        InstructNav \cite{long2024instructnav}
        & -- & \cmark & \cmark & \cmark &
        & 6.89 & -- & 31.0 & 24.0
        & -- & -- & -- & -- & -- \\

        \midrule

        COSMO \cite{zhang2025cosmo}
        & -- & \cmark & & &
        & -- & 56.0 & 47.0 & 40.0
        & -- & -- & -- & -- & -- \\

        AO-Planner \cite{chen2025affordances}
        & -- & \cmark & & \cmark &
        & 5.55 & 59.0 & 47.0 & 33.0
        & 7.06 & 43.3 & 30.5 & 50.1 & -- \\

        LAW \cite{raychaudhuri2021language}
        & -- & & \cmark & \cmark & \cmark
        & 6.83 & 44.0 & 35.0 & 31.0
        & 10.90 & 8.0 & 8.0 & 38.0 & -- \\

        MapNav \cite{zhang2025mapnav}
        & -- & & \cmark & \cmark & \cmark
        & \underline{4.93} & 53.0 & 39.7 & 37.2
        & -- & -- & -- & -- & -- \\

        g3D-LF \cite{wang2025g3d}
        & -- & & \cmark & \cmark & \cmark
        & 5.70 & 59.5 & 47.2 & 34.6
        & -- & -- & -- & -- & -- \\

        Seq2Seq \cite{krantz2020beyond}
        & -- & & & \cmark & \cmark
        & 7.77 & 37.0 & 25.0 & 22.0
        & 12.10 & 13.9 & 11.9 & 30.8 & -- \\

        NaVid-4D \cite{liu2025vid}
        & -- & & & \cmark & \cmark
        & 5.99 & 55.7 & 43.8 & 37.1
        & -- & -- & -- & -- & -- \\

        NavMorph \cite{yao2025navmorph}
        & -- & & & \cmark & \cmark
        & 5.75 & 56.9 & 47.9 & 33.2
        & 8.85 & 30.8 & 22.8 & 44.2 & -- \\

        \midrule

        NaVid \cite{zhang2024navid}
        & 7B & & & & \cmark
        & 5.47 & 49.1 & 37.4 & 35.9
        & -- & -- & -- & -- & 953K \\

        Sim2Real \cite{wang2024sim}
        & -- & & & & \cmark
        & 5.95 & 55.8 & 44.9 & 30.4
        & 8.79 & 36.7 & 25.5 & 18.1 & 0K \\

        Uni-NaVid \cite{zhang2024uni}
        & 7B & & & & \cmark
        & 5.58 & 53.3 & 47.0 & 42.7
        & 6.24 & 48.7 & 40.9 & -- & 3577K \\

        InternVLA-N1 \cite{wei2026ground}
        & 8B & & & & \cmark
        & \textbf{4.89}
        & 60.6
        & \underline{55.4}
        & \textbf{52.1}
        & 6.41
        & 49.5
        & 41.8
        & \underline{62.6}
        & 51200K \\

        StreamVLN$^{*}$ \cite{wei2025streamvln}
        & 7B & & & & \cmark
        & 6.05 & 53.8 & 45.5 & 41.6
        & -- & -- & -- & -- & 10033K \\

        NaVILA$^{*}$ \cite{cheng2024navila}
        & 7B & & & & \cmark
        & 5.22
        & \underline{62.5}
        & 54.0
        & 49.0
        & 6.77
        & 49.3
        & \underline{44.0}
        & 58.8
        & 13132K \\

        Aux-Think \cite{wang2026think}
        & 8B & & & & \cmark
        & 6.08
        & 60.0
        & 54.8
        & 46.9
        & 6.24
        & \underline{52.2}
        & 40.2
        & --
        & 1600K \\

        ActiveVLN \cite{zhang2025activevln}
        & 3B & & & & \cmark
        & 5.31
        & 58.3
        & 50.1
        & 43.7
        & \underline{5.84}
        & 50.7
        & 41.2
        & 58.1
        & 0K \\

        JanusVLN$^{\dagger}$ \cite{zeng2026janusvln}
        & 7B + 1B & & & & \cmark
        & 5.17
        & 58.0
        & 52.8
        & 49.2
        & 6.46
        & 51.4
        & \textbf{44.3}
        & 59.1
        & 0K \\

        \rowcolor{blue!7}
        StageVLN (Ours)
        & 4B & & & & \cmark
        & 5.17
        & \textbf{64.7}
        & \textbf{56.3}
        & \underline{51.4}
        & \textbf{5.65}
        & \textbf{54.3}
        & 42.8
        & 58.3
        & 0K \\

        \bottomrule
    \end{tabular}%
    }
\end{table}

\begin{table}[t]
    \centering
    \caption{Inference memory (peak allocated VRAM) and mean step time,
    grouped by GPU. StageVLN is shaded; bold values indicate the best
    result within each group. OOM denotes an out-of-memory failure
    under the tested configuration.}
    \label{tab:stagevln_efficiency}
    \small
    \renewcommand{\arraystretch}{1.15}
    \setlength{\tabcolsep}{8pt}
    \begin{tabular}{lrrrr}
        \toprule
        \multirow{2}{*}{Method} &
        \multicolumn{2}{c}{RTX 5090} &
        \multicolumn{2}{c}{RTX 6000 Pro} \\
        \cmidrule(lr){2-3}
        \cmidrule(lr){4-5}
        & VRAM (GiB)$\downarrow$ & Step (ms)$\downarrow$
        & VRAM (GiB)$\downarrow$ & Step (ms)$\downarrow$ \\
        \midrule
        JanusVLN  & OOM & -- & 35.73 & 481.3 \\
        NaVILA    & 17.47 & 365.2 & 17.47 & 383.1 \\
        StreamVLN & 17.48 & \textbf{233.4} & 17.48 & \textbf{209.8} \\
        \rowcolor{blue!7}
        \textbf{StageVLN} & \textbf{10.24} & 304.5 & \textbf{10.24} & 229.8 \\
        \bottomrule
    \end{tabular}
\end{table}

\subsubsection{Implementation Details}
We instantiate StageVLN with Qwen3.5-4B~\cite{qwen3.5} and a frozen
VGGT-$\Omega$~\cite{wang2026vggt} geometry teacher. The navigator receives up to
eight historical RGB frames and the current observation,
while the teacher processes only the current frame.
Navigator hidden states after layers 8, 16, and 24 are
aligned with VGGT-$\Omega$ aggregator outputs at zero-based indices
11, 17, and 23, respectively. Each alignment projector uses LayerNorm
followed by a two-layer MLP with dimensions
$2560\rightarrow4096\rightarrow2048$ and GELU activation.
We set $\ell_h=\ell_p=24$, with relative-heading targets computed
using nominal turn increments of $\pm15^\circ$.

We set $\lambda_s=0.3$ and $\lambda_h=\lambda_p=0.1$,
averaging the spatial loss equally across the three levels.
Training uses AdamW with learning rates of $10^{-6}$ for
the language model and its output head, and $10^{-5}$ for
the visual merger, alignment projectors, and trajectory heads.
The visual encoder and geometry teacher remain frozen.
We train for one epoch with a global batch size of 64 using four NVIDIA H100 GPUs.

\begin{figure}[t]
    \centering
    \includegraphics[width=\textwidth]{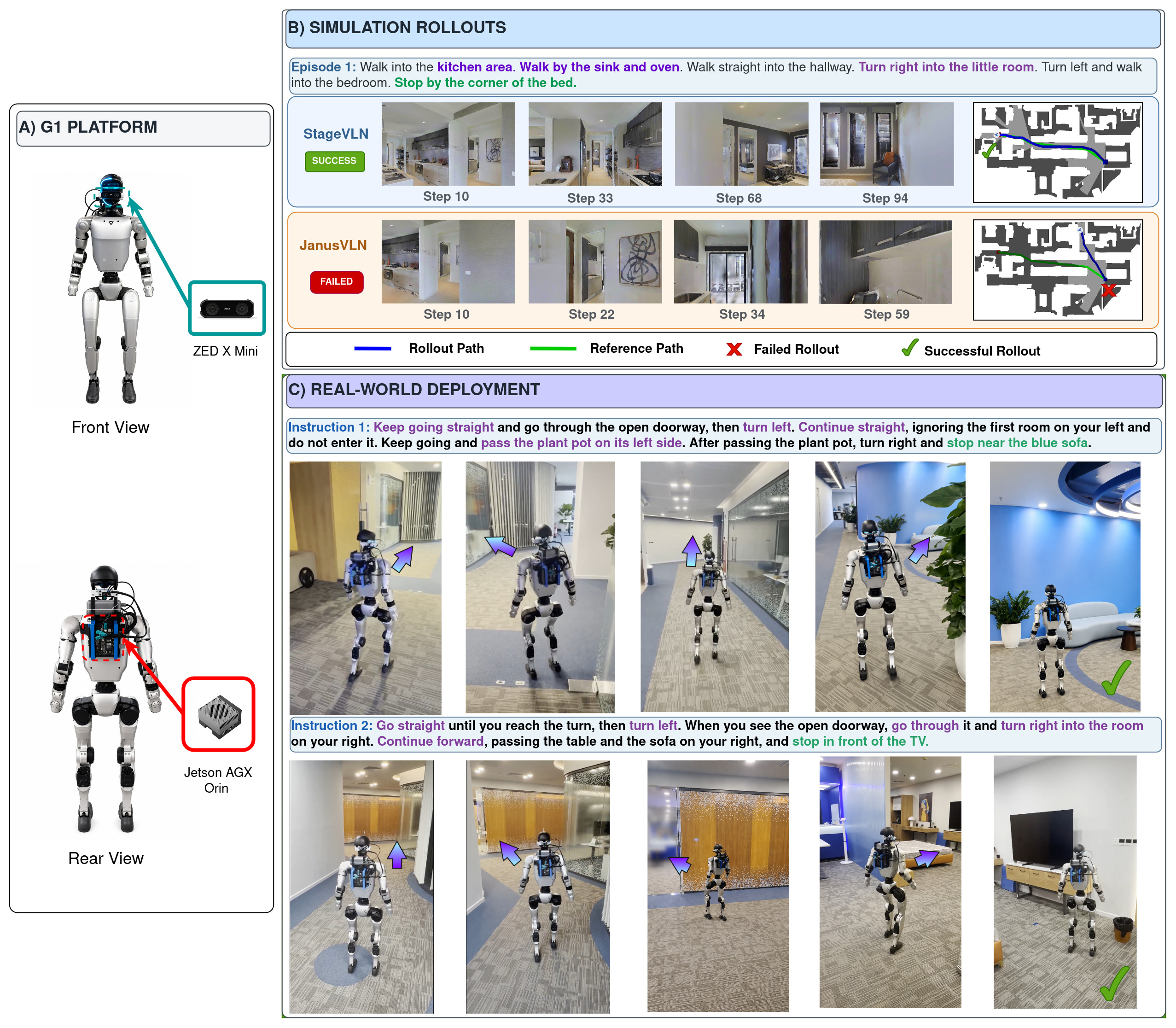}
    \caption{\textbf{Simulation rollouts and real-world deployment.}
    (a) Unitree G1 platform with a ZED X Mini camera and Jetson AGX Orin.
    (b) Simulated episode comparing JanusVLN and StageVLN
    under the same instruction and initial condition.
    (c) Two real-world deployment sequences, shown in
    chronological order.}
    \label{fig:stagevln_rollouts}
\end{figure}

\subsubsection{Real-World Setup}
We deploy StageVLN on a Unitree G1 humanoid equipped with
a ZED X Mini camera (Fig.~\ref{fig:stagevln_rollouts}).
The Jetson AGX Orin handles image preprocessing and communication
with a remote workstation equipped with an NVIDIA GeForce RTX 5090,
where policy inference is performed. The geometry teacher, alignment
projectors, and trajectory heads are excluded from deployment; only
RGB observations and language instructions are provided to the
StageVLN policy.

\subsection{Main Benchmark Results}
Table~\ref{tab:stagevln_main_results} reports results on the
R2R-CE and RxR-CE validation-unseen splits. With a 4B navigator,
StageVLN achieves SR/SPL of 56.3/51.4 on R2R-CE and 54.3/42.8
on RxR-CE, without additional navigation training data or an
extra geometry encoder at inference.

On R2R-CE, StageVLN achieves the highest OS (64.7) and SR (56.3)
among the listed methods, together with the second-best SPL (51.4).
On RxR-CE, it obtains the highest SR (54.3) and lowest NE (5.65).
Notably, despite using a 4B navigator, StageVLN achieves
higher SR on both benchmarks than several larger 7B--8B models,
including InternVLA-N1~\cite{wei2026ground}, NaVILA~\cite{cheng2024navila}, and JanusVLN~\cite{zeng2026janusvln}.

Compared with JanusVLN~\cite{zeng2026janusvln}, which also
leverages geometric representations but retains an additional
1B geometry encoder at inference, StageVLN uses geometry
supervision only during training and improves SR by 3.5 and
2.9 percentage points on R2R-CE and RxR-CE, respectively.
R2R-CE SPL also improves by 2.2 points, whereas RxR-CE SPL is
1.5 points lower. These results suggest that training-time geometric supervision can
effectively enhance goal-reaching performance without augmenting the inference architecture with a separate geometry encoder.

\subsection{Deployment Efficiency}
Table~\ref{tab:stagevln_efficiency} compares memory usage
and mean step time within each GPU group. For the baselines, we
follow the released code and recommended configurations, and report the resulting measurements under our
hardware setup. In the reported
runs, StageVLN uses 41.4\% less peak allocated VRAM than
NaVILA~\cite{cheng2024navila} on both devices, with step times reduced by 16.6\%
on the RTX 5090 and 40.0\% on the RTX 6000 Pro.
StreamVLN~\cite{wei2025streamvln} remains faster, highlighting a memory--latency
trade-off. JanusVLN~\cite{zeng2026janusvln} encounters OOM on the RTX 5090 under
the tested configuration.

\subsection{Ablation Studies}

\subsubsection{Component Ablation}
Table~\ref{tab:stagevln_components} examines spatial and
trajectory guidance on R2R-CE validation-unseen. All controlled
ablations use VGGT~\cite{wang2025vggt}, whereas the headline StageVLN in
Table~\ref{tab:stagevln_main_results} uses VGGT-$\Omega$~\cite{wang2026vggt}.
Multi-level geometric alignment outperforms single-level alignment
(52.26/46.72 versus 48.50/41.86 SR/SPL), while adding heading
supervision alone reduces SR/SPL. Trajectory supervision without
geometric distillation already achieves 54.43/48.58 SR/SPL.
Adding multi-level geometric alignment further improves SR to 55.36,
reduces NE from 5.86 to 5.58, and increases OS from 64.06 to 67.37,
while SPL remains similar (48.44). Compared with SFT, the full
formulation improves SR/SPL by 8.87/6.12 points, although NE remains
higher (5.58 versus 5.19). Overall, the objectives provide
complementary benefits without uniformly improving every metric.

\begin{table}[t]
    \centering
    \small
    \begin{minipage}[t]{0.59\linewidth}
    \centering
    \caption{Component ablation on R2R-CE validation-unseen. VGGT is used as the geometry teacher for all variants.
    \textbf{Nav.}: navigation supervision; \textbf{Geo.}: geometric
    representation distillation; \textbf{Head.}: relative heading;
    \textbf{Prog.}: expert-route progress.
    1L uses only the final alignment pair (navigator layer 24,
    teacher index 23), while ML uses all three alignment pairs.
    The first row is supervised fine-tuning (SFT); the shaded row is
    full StageVLN. Best results are shown in \textbf{bold}.}
    \label{tab:stagevln_components}
    \setlength{\tabcolsep}{3.5pt}
    \renewcommand{\arraystretch}{1.12}
    \begin{tabular}{ccccrrrr}
        \toprule
        \textbf{Nav.} & \textbf{Geo.} & \textbf{Head.} & \textbf{Prog.} &
        NE$\downarrow$ & OS$\uparrow$ &
        SR$\uparrow$ & SPL$\uparrow$ \\
        \midrule

        \cmark & -- & & &
        \textbf{5.19} & 56.01 & 46.49 & 42.32 \\

        \cmark & 1L & & &
        6.33 & 61.83 & 48.50 & 41.86 \\

        \cmark & ML & & &
        6.12 & 61.01 & 52.26 & 46.72 \\

        \cmark & ML & \cmark & &
        6.17 & 63.30 & 49.97 & 43.80 \\

        \cmark & ML & & \cmark &
        5.66 & 63.89 & 53.18 & \textbf{48.67} \\

        \cmark & -- & \cmark & \cmark &
        5.86 & 64.06 & 54.43 & 48.58 \\

        \rowcolor{blue!6}
        \cmark & ML & \cmark & \cmark &
        5.58 & \textbf{67.37} & \textbf{55.36} & 48.44 \\

        \bottomrule
    \end{tabular}
    \end{minipage}\hfill
    \begin{minipage}[t]{0.37\linewidth}
    \centering
    \caption{Geometry-teacher ablation on R2R-CE
    validation-unseen. Each variant uses the full StageVLN
    objective with a frozen teacher. Best results are shown in
    \textbf{bold}.}
    \label{tab:stagevln_teachers}
    \setlength{\tabcolsep}{3.5pt}
    \renewcommand{\arraystretch}{1.12}
    \begin{tabular}{lrrrr}
        \toprule
        Teacher &
        NE$\downarrow$ & OS$\uparrow$ &
        SR$\uparrow$ & SPL$\uparrow$ \\
        \midrule
        VGGT
        & 5.58
        & \textbf{67.37}
        & 55.36
        & 48.44 \\
        VGGT-$\Omega$
        & \textbf{5.17}
        & 64.71
        & \textbf{56.28}
        & \textbf{51.37} \\
        \bottomrule
    \end{tabular}
    \end{minipage}
\end{table}

\begin{figure}[p]
    \centering
    \setlength{\rwgap}{3pt}
    \setlength{\rwwidth}{\dimexpr(\linewidth-4\rwgap)/5\relax}

    \rwinstr{Instruction 3}{\rwact{Go straight} until you reach the
    \rwact{white cabinet}. \rwact{Move to the right side of the cabinet}
    to pass it, then \rwact{go through the open doorway}.
    \rwgoal{Stop near the green plant pot.}}\par\vspace{2pt}
    \rwframe{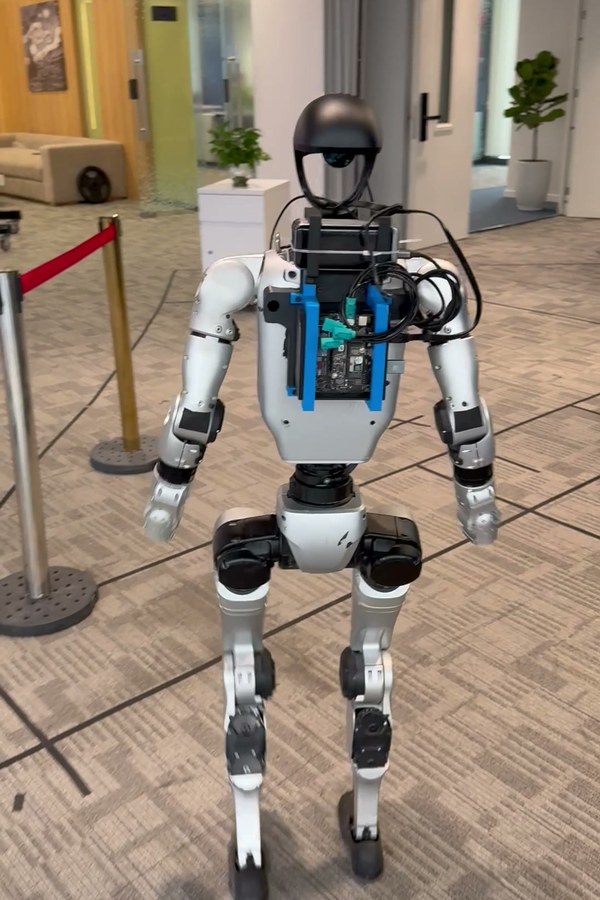}{10\,s}{0}\hfill\rwframe{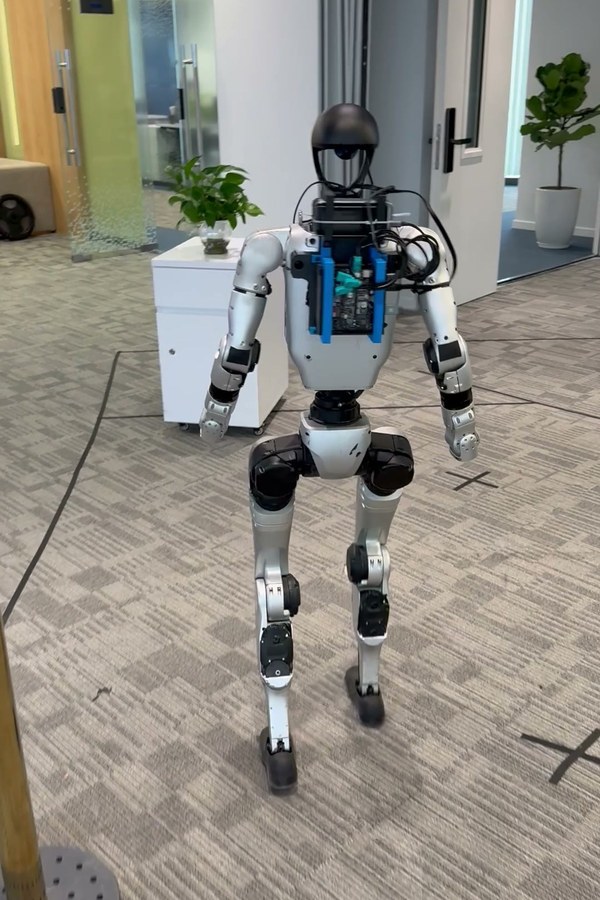}{17\,s}{0}\hfill%
    \rwframe{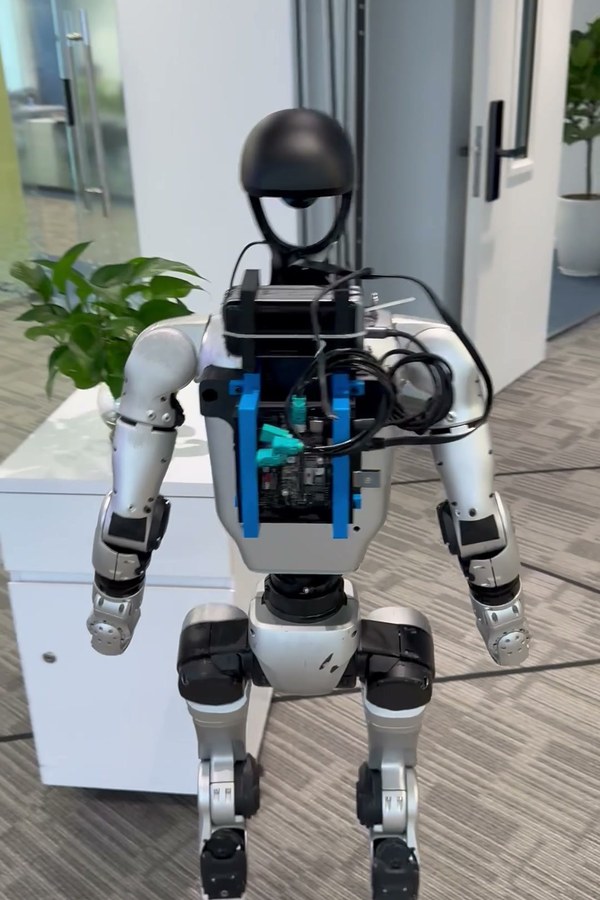}{22\,s}{0}\hfill\rwframe{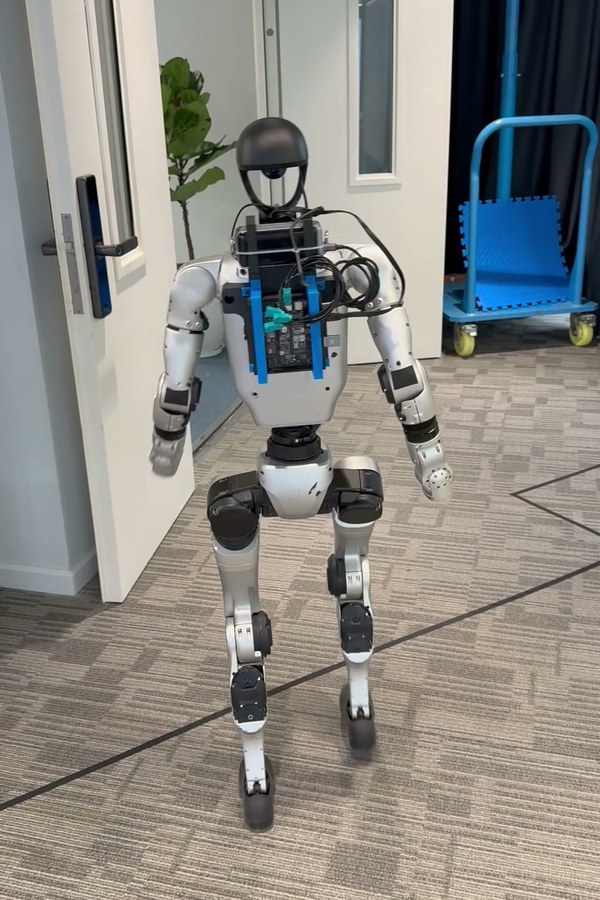}{29\,s}{0}\hfill%
    \rwframe{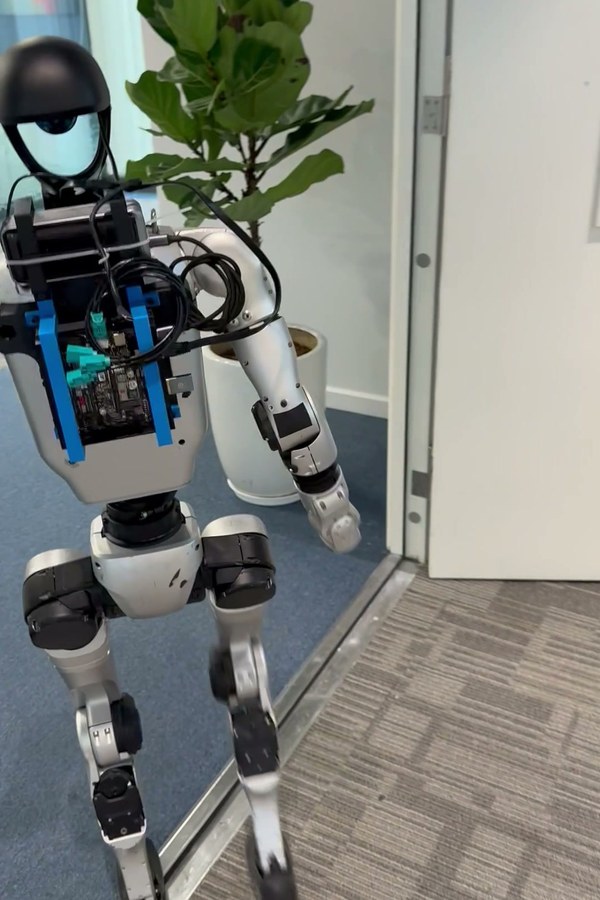}{37\,s}{1}\par\vspace{6pt}

    \rwinstr{Instruction 4}{\rwact{Go straight then turn right.}
    Continue straight until you reach the \rwact{intersection}, then
    \rwact{veer slightly to the right}. \rwgoal{Stop at the position
    between the green plant pot and the fire extinguisher box.}}\par\vspace{2pt}
    \rwframe{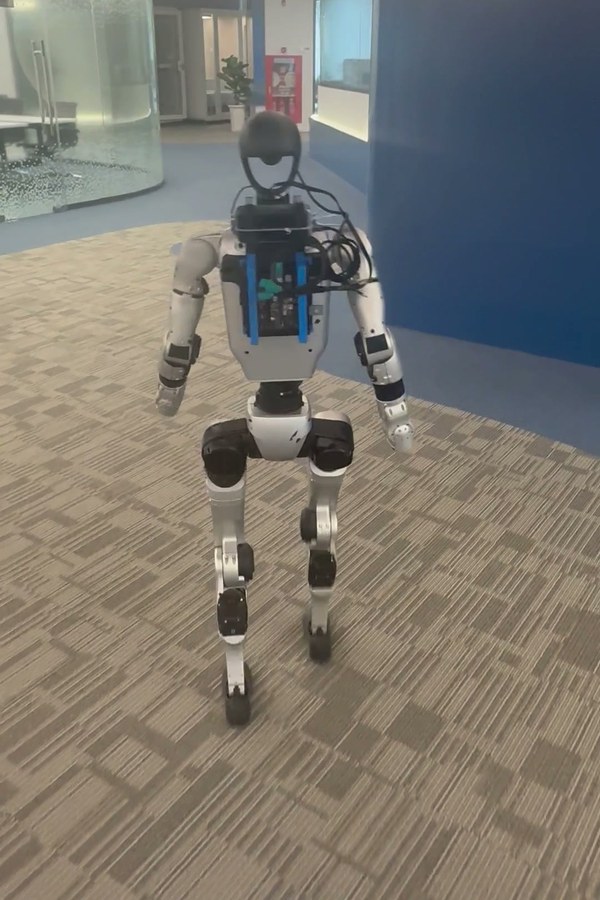}{12\,s}{0}\hfill\rwframe{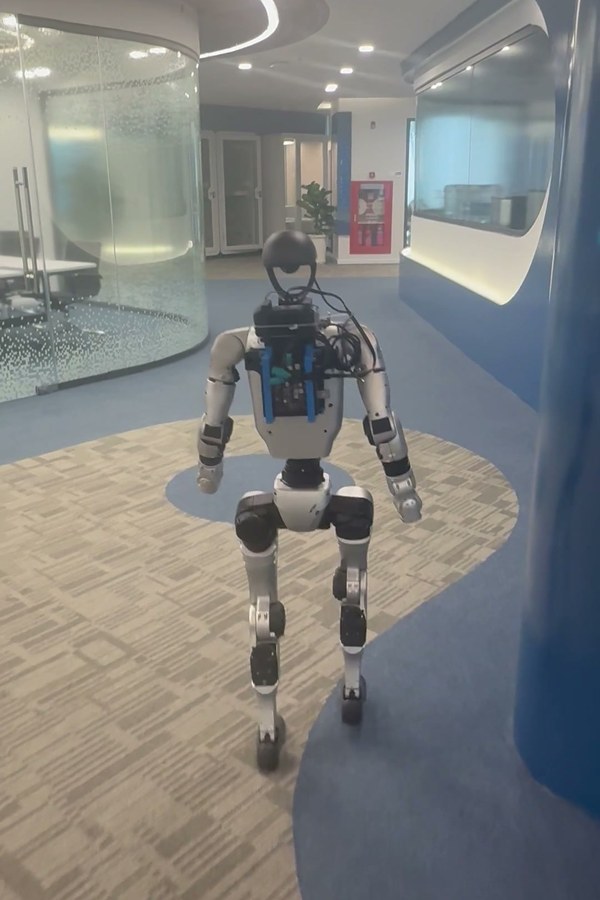}{20\,s}{0}\hfill%
    \rwframe{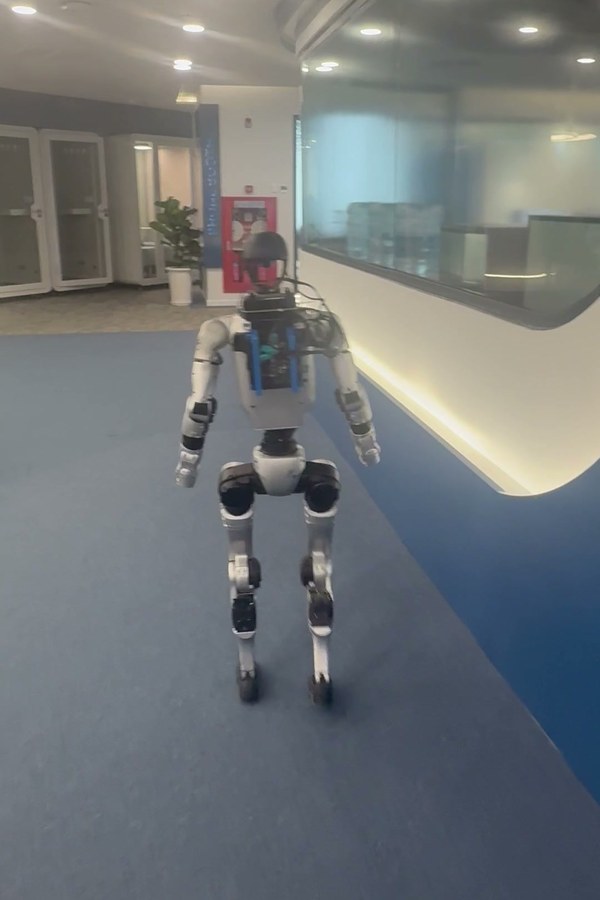}{33\,s}{0}\hfill\rwframe{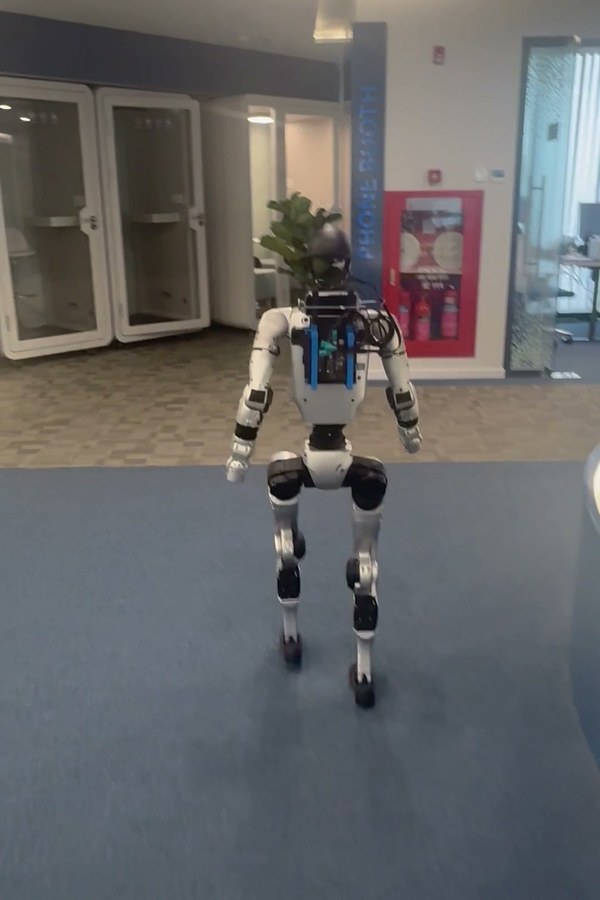}{46\,s}{0}\hfill%
    \rwframe{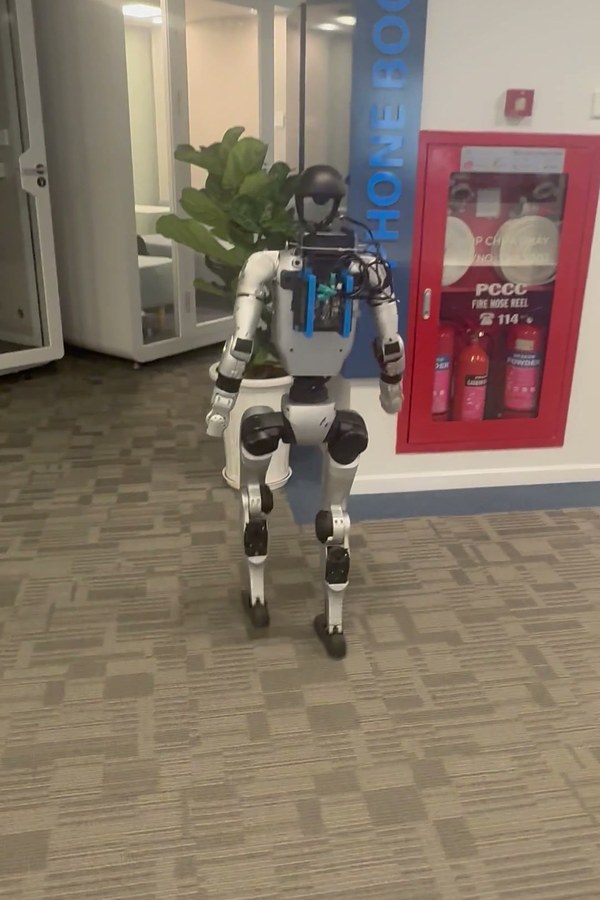}{54\,s}{1}\par\vspace{6pt}

    \rwinstr{Instruction 5}{\rwact{Go straight, passing the plant on its
    left side.} Continue straight until you reach the
    \rwact{intersection}, then \rwact{turn left}. Keep going straight
    and \rwgoal{stop near the plant at the end of the hallway.}}\par\vspace{2pt}
    \rwframe{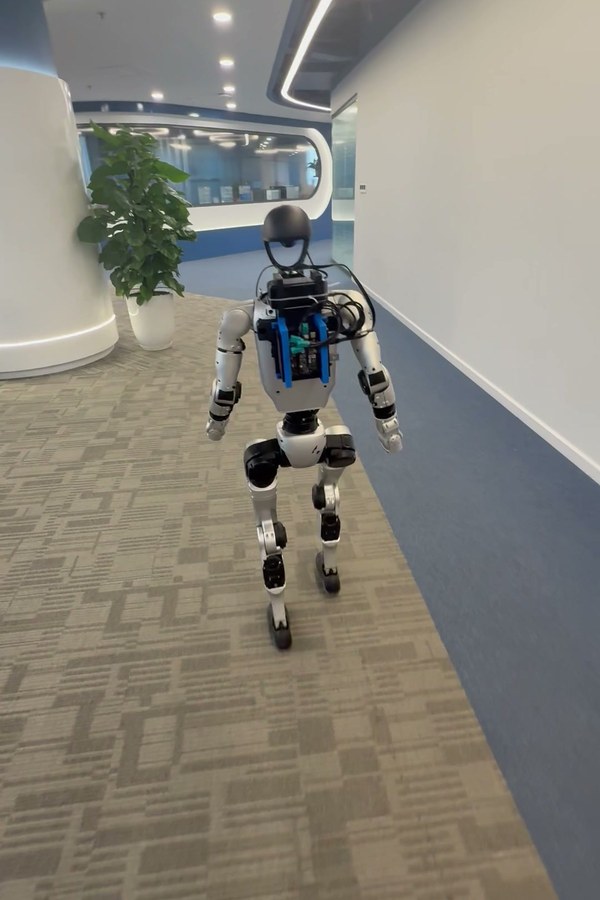}{10\,s}{0}\hfill\rwframe{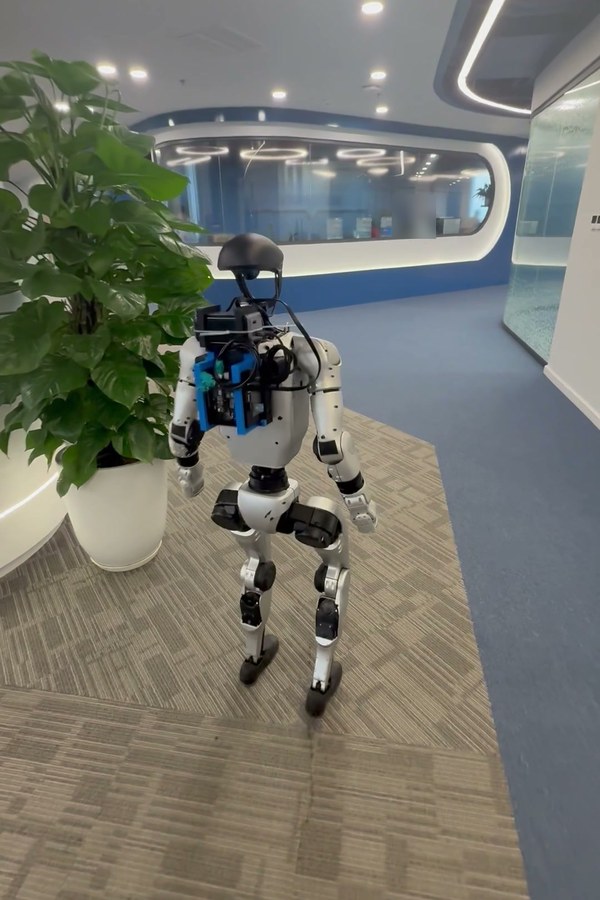}{18\,s}{0}\hfill%
    \rwframe{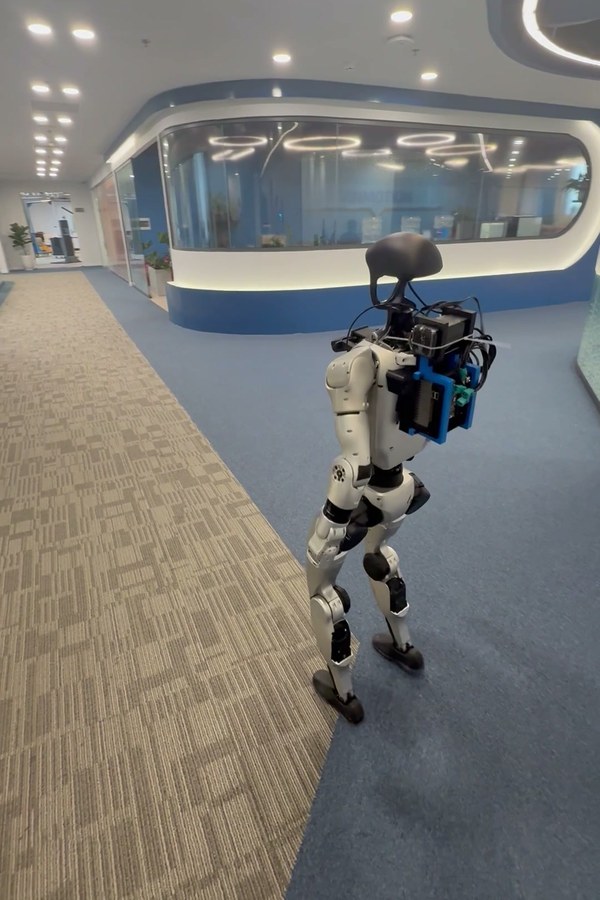}{32\,s}{0}\hfill\rwframe{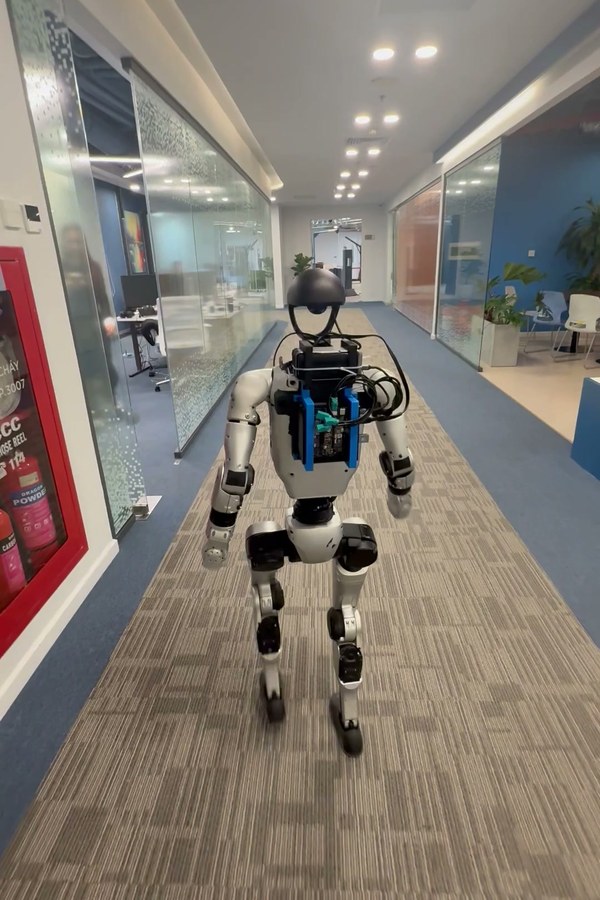}{45\,s}{0}\hfill%
    \rwframe{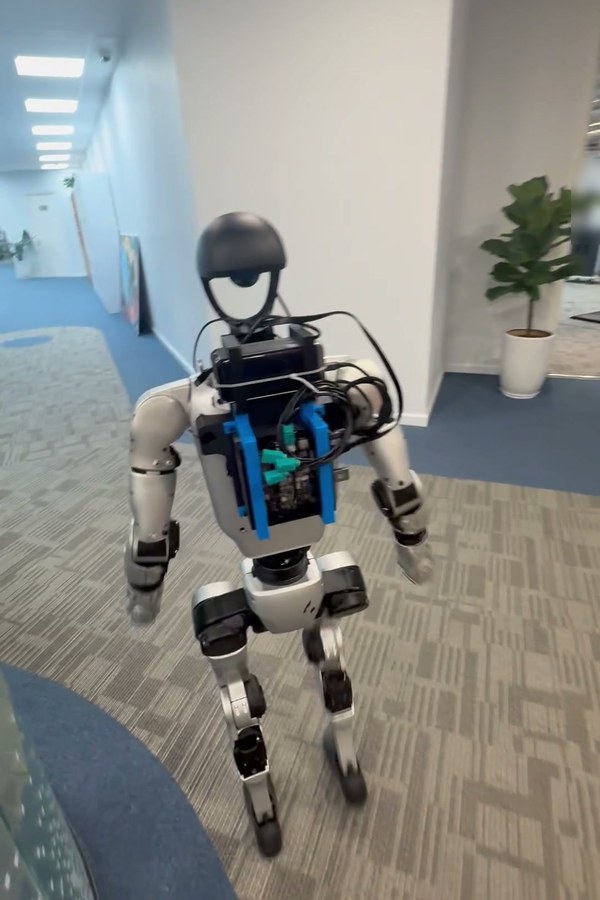}{82\,s}{1}

    \caption{\textbf{Additional real-world episodes.} Three more
    instructions executed by StageVLN on the Unitree G1 without
    real-world fine-tuning. Numbering continues from the two
    instructions in Fig.~\ref{fig:stagevln_rollouts}(c). Third-person frames are in
    chronological order; each label gives the time since the start
    of the recording. Purple marks route steps and landmarks, green marks the goal,
    and \ding{51} marks the final stop.}
    \label{fig:stagevln_realworld}
\end{figure}

\subsubsection{Geometry-Teacher Ablation}

To examine how the choice of geometry teacher affects StageVLN
performance, we use VGGT~\cite{wang2025vggt} and VGGT-$\Omega$~\cite{wang2026vggt} during training under the same
navigation setup. As shown in Table~\ref{tab:stagevln_teachers},
VGGT-$\Omega$ achieves the best NE, SR, and SPL
(5.17/56.28/51.37), while VGGT yields the highest OS (67.37).
These results indicate that teacher choice and representation
quality affect downstream navigation performance.

\subsection{Qualitative Analysis and Real-World Deployment}

Figure~\ref{fig:stagevln_rollouts}(b) qualitatively compares
StageVLN with JanusVLN~\cite{zeng2026janusvln} under the same
instruction and initial condition. In the illustrated episode,
StageVLN follows the reference route and successfully reaches
the target, whereas JanusVLN deviates from the intended route
early in the rollout and consequently fails to complete the
instruction.

We further demonstrate StageVLN on the real-world setup described
above. As shown in Fig.~\ref{fig:stagevln_rollouts}(c), StageVLN
successfully completes both real-world navigation instructions,
following multi-step routes involving corridor traversal,
doorway transitions, and object-referenced navigation, while
stopping at locations that closely match the instructed goals.

Figure~\ref{fig:stagevln_realworld} shows three additional indoor
episodes. Their instructions combine passing a
landmark on a specified side (the white cabinet in Instruction~3 and
the potted plant in Instruction~5), turns at intersections, including
a gradual one (``veer slightly to the right'' in Instruction~4), and
goals defined relative to nearby landmarks: a plant pot reached through
an open doorway, the position between a plant pot and a fire extinguisher
box, and a plant at the end of a long hallway. In all three episodes,
StageVLN follows the described route and stops at the instructed goal.
No additional fine-tuning or adaptation on real-world navigation data
is applied in any of the five episodes, demonstrating the feasibility
of transferring StageVLN from simulation to a physical robot. These
episodes are qualitative examples rather than a systematic success-rate
evaluation.

\section{Conclusion}
We presented StageVLN, a framework for training
vision--language navigation policies with complementary
spatial and trajectory guidance. Multi-level geometric
representation distillation supervises hierarchical scene
structure, while relative-heading and expert-route progress
objectives encourage relational orientation and global
trajectory-state representations. All auxiliary modules
are used only during training, preserving the base
RGB--language inference pathway. Across the evaluated settings, the results support training-time spatial and trajectory guidance as a way to improve success-related navigation metrics without requiring a runtime geometry encoder.

\bibliographystyle{unsrtnat}
\bibliography{references}

\end{document}